# Adversarial Multiscale Feature Learning for Overlapping Chromosome Segmentation

Liye Mei, *Student Member*, *IEEE*, Yalan Yu, Yueyun Weng, Xiaopeng Guo, *Student Member*, *IEEE*, Yan Liu, Du Wang, Sheng Liu, *Fellow*, *IEEE*, Fuling Zhou, and Cheng Lei

*Abstract*—**Chromosome karyotype analysis is of great clinical importance in the diagnosis and treatment of diseases, especially for genetic diseases. Since manual analysis is highly time and effort consuming, computer-assisted automatic chromosome karyotype analysis based on images is routinely used to improve the efficiency and accuracy of the analysis. Due to the strip shape of the chromosomes, they easily get overlapped with each other when imaged, significantly affecting the accuracy of the analysis afterward. Conventional overlapping chromosome segmentation methods are usually based on manually tagged features, hence, the performance of which is easily affected by the quality, such as resolution and brightness, of the images. To address the problem, in this paper, we present an adversarial multiscale feature learning framework to improve the accuracy and adaptability of overlapping chromosome segmentation. Specifically, we first adopt the nested U-shape network with dense skip connections as the generator to explore the optimal representation of the chromosome images by exploiting multiscale features. Then we use the conditional generative adversarial network (cGAN) to generate images similar to the original ones, the training stability of which is enhanced by applying the least-square GAN objective. Finally, we employ Lovász-Softmax to help the model converge in a continuous optimization setting. Comparing with the established algorithms, the performance of our framework is proven superior by using public datasets in eight evaluation criteria, showing its great potential in overlapping chromosome segmentation.**

## I. INTRODUCTION

HUMAN chromosome karyotype analysis is of great diagnostic and prognostic value in diseases. It is usually performed in clinical diagnosis, cancer cytogenetic, and genetic abnormalities detecting such as Edwards syndrome and Down syndrome [1, 2]. The morphology of chromosomes, such as the excess or missing, or the structural defects of specific chromosomes can be directly linked to corresponding diseases, hence chromosome karyotype analysis based on images plays a critical role in routine disease diagnosis and treatment [3]. Fig. 1 illustrates the process of chromosome karyotype analysis, which consists of two main steps: *segmentation* and *classification*. The performance of the segmentation can directly influence the accuracy of the classification afterward. As shown in the red circles in Fig. 1(a) and Fig. 1 (b), the overlapping chromosome segmentation is even more challenging due to the ambiguity in the overlapping regions, which can greatly influence the accuracy of the chromosome karyotype analysis. Hence, segmenting overlapping chromosomes with high efficiency and accuracy can significantly enhance the overall performance of the chromosome karyotype analysis.

Since manual segmentation is both time and effort consuming, and the accuracy highly depends on the experiences of the analyst, over the past few decades, many algorithms have been proposed to automatically segment the chromosome from the images on computers. According to their principles, these segmentation methods can be roughly classified into two categories: heuristic methods and learning-based methods. For heuristic methods, they utilize manually tagged features, such as contour, pixels, geometric features, to perform segmentation [4-6]. The representative examples include Ritter et al. utilized shape analysis and classification for chromosome segmentation and adopt global context and variant analysis method to solve complex and ambiguous cases [7]. However, it consists of two phases and somewhat cumbersome. Madian et al. used the contour analysis method and constructing reasonable hypotheses for segmentation and separation [8]. Saiyod et al. proposed an edge detection method, which consists of the flood-fill, Erosion, and Canny method [9], however it only solve the touching chromosome, not the overlapping chromosome. Some researchers usually use thresholding strategies for chromosome segmentation [10-12], adopting local adaptive histogram equalization technique to obtain the appropriate threshold, to further enhance chromosome

This research was supported by the National Natural Science Foundation of China under contracts 61905182 and 51727901; 2020 Medical Science and Technology Innovation Platform Support Project of Zhongnan Hospital of Wuhan University; Wuhan Research Program of Application Foundation and Advanced Technology (Corresponding author: Cheng Lei)

L. Mei and Y. Yu contributed equally to this work.

L. Mei, D. Wang, S. Liu and C. Lei are with the Institute of Technological Sciences, Wuhan University, Wuhan, 430072, China (e-mail: liye_mei@outlook.com; wangdu@whu.edu.cn; victor_liu63@vip.126.com; leicheng@whu.edu.cn)

Y. Yu and F. Zhou are with the Department of Hematology, Zhongnan Hospital of Wuhan University, Wuhan, 430071, China (e-mail: yuyalan@znhospital.cn; zhoufuling@whu.edu.cn)

Y. Weng is with the School of Power and Mechanical Engineering, Wuhan University, Wuhan, 430072, China (e-mail: wengyueyun@whu.edu.cn)

X. Guo is with the Wangxuan Institute of Computer Technology, Peking University, Beijing 100080, China (email: xiaopeng.guo@stu.pku.edu.cn)

Y. Liu is with the Alipay Tian Qian Security Lab, Beijing,100020, China (e-mail: Goodmandou@gmail.com).

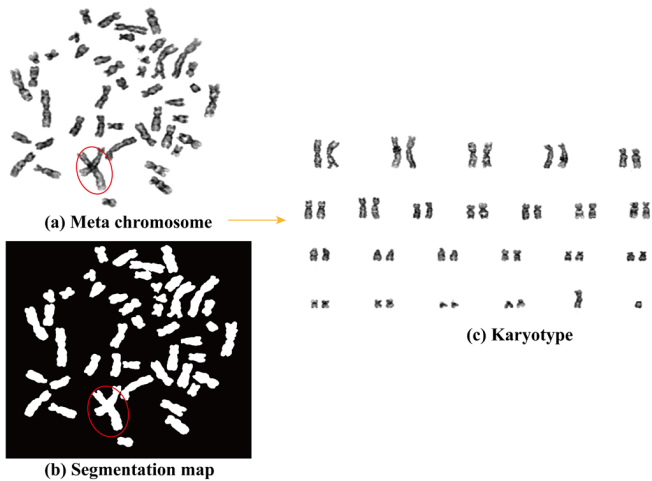

Fig.1. Chromosome karyotype analysis process. (a) The image of meta chromosome. (b) Segmentation map of the chromosome image. (c) The sorted karyotype. The red circles make the overlapping regions.

segments by reducing the chances of pixel misclassification. But this method is susceptible to noise, contrast and resolution in the image. Gawande et al. applied a fuzzy C-means clustering algorithm and watershed algorithm for chromosome segmentation, but it also didn't effectively separate overlapping chromosomes [13]. Sharma et al. adopt a combination of crowdsourcing for segmentation, but it required considerable effort and time to separate chromosomes manually [14]. Lin et al. proposed a geometric feature to separate chromosomes [15], however, it can't automate chromosome segmentation well due to the irregular shape of chromosomes. These methods can achieve impressive segmentation results when manual features are properly tagged, however, they are very sensitive to the shape and overlapping regions of the chromosomes. Moreover, since they do not consider the untagged features, the performance and applicability of these methods are limited, and it is difficult to implement on large data sets. While, for learning-based methods, they usually apply machine learning techniques to mine potential information from the images to perform chromosome segmentation and medical image analysis [16, 17]. Some representative examples include Pardo et al. applied fully convolution network (FCN) method for Karyotype analysis [18, 19]. However it required multicolor in situ hybridization technique to obtain the original image, where there were no overlapping chromosomes, it therefore is difficult to meet the clinical reality. Other researchers use a U-shape Network (UNet) for overlapping chromosome segmentation [20, 21]. Chen et al. proposed the shape learning method to segment both non-overlapped and overlapped regions [22]. Altinsoy et al. proposed a raw G-band chromosome image segmentation method using convolution network [23], but it didn't work for overlapping chromosomes. These methods can independently conduct chromosome segmentation when being trained. However, limited by the architecture of the network, current learning-based methods only utilize several layers' features and they do not take advantage of multiscale features to adapt different chromosome scales. Hence, they do not perform well in overlapping chromosome segmentation.

In this paper, considered the various scales and overlapping regions of chromosome images, we demonstrate an adversarial multiscale feature learning (AMFL) framework which employs nested U-shape convolutional neural network (NestedUNet) [24], conditional generative adversarial network (cGAN) [25], and Lovász-Softmax [26] for overlapping chromosome segmentation. Specifically, NestedUNet consists of UNets [27] of varying depths and owns dense skip connectivity, making it capable of synthesizing multiscale feature maps for segmentation. Hence, our AMFL framework utilizes NestedUNet to explore the optimal representation of chromosome images by exploiting multiscale features and features fused. Meanwhile, we consider the overlapping chromosome image segmentation as an images-to-image task, in which the source overlapping chromosome images are translated to a confidence map to indicate the category information in the source images, we therefor use cGAN to push the output distributions close to the ground truths for its success in computer vision tasks, such as blind motion deblurring[28], image fusion [29], and image inpainting [30]. Finally, to optimize the performance of the discriminatively trained overlapping chromosome segmentation, we apply Lovász-Softmax, which is based on the convex lovász extension of the submodular loss, as the segmentation loss to achieve superior chromosome segmentation performance and higher index scores to the traditional cross-entropy (CE) loss. Additionally, we utilize the least-square GAN objective [31] to replace the original GAN loss in the overlapping chromosome segmentation task to stabilize the training and avoid model collapsing. To verify the feasibility of our method, we carry out extensive experiments to compare the performance of our AMFL framework with others. Results show the superiority of our AMFL framework in this work in terms of visual perception analysis and quantitative score comparison.

## II. METHODS

In this section, we introduce the architecture of our AMFL framework, the quantitative criteria to evaluate the performance of the chromosome segmentation, and the references which are used to validate the feasibility of our method.

### A. Network architecture

As depicted in Fig. 2, our AMFL framework consists of two modules: a generator and a discriminator. The generator is responsible for exploiting multiscale features for segmentation by producing "fake" chromosome images. While the discriminator is to distinguish the "fake" images from the "real" ones by adversarial learning. Once the discriminator is "fooled" by the generator, the network is ready to segment chromosomes with high accuracy.

*1) Generator* As shown in Fig. 2(a), we adopt the advanced NestedUNet [24] as the generator G, which consists of an encoder and a decoder. It takes a source chromosome image as the input and outputs a multiclass one-hot map. Specifically, each node in the graph represents a nested convolution block, the downward arrows, upward arrows, and dotted arrows indicate $2 \times 2$ max-pooling, $2 \times 2$ up-sampling, and skip connections, respectively. The skip connections merge the encoding and decoding features in the channel dimension by tensor concatenation，enabling dense feature propagation. To better comprehend the network connectivity, we denote $x^{i,j}$ the output of the node $X^{i,j}$. It can be formulated as:

$$x^{i,j} = \begin{cases} \mathcal{N}(\mathcal{D}(x^{i-1,j})), & j = 0 \\ \mathcal{N}([x^{i,k}]_{k=0}^{j-1}, \mathcal{U}(x^{i+1,j} - 1)), & j > 0 \end{cases} \quad (1)$$

where function $\mathcal{N}(\cdot)$ denotes a Nested convolution block operation, $\mathcal{D}(\cdot)$ and $\mathcal{U}(\cdot)$ denote a down-sampling layer and an up-sampling layer respectively, and $[\cdot]$ denotes the concatenation layer. Intuitionly, we can see that the nodes at the level of $j = 0$ receive only one input from the previous layer of the encoder, whereas the nodes at the level of $j > 1$ receive the up-sampled output of $j + 1$ nodes from the lower skip connection and all the outputs of the previous $j$ nodes in the same skip connection. Therefore, a dense skip connection is constructed and multiscale features are integrated to provide better feature representation for the segmentation of overlapping chromosome regions with different scales. Meanwhile, in order to better describe network parameters, the number of filters is defined as: $f = \{64,128,256,512,1024\}$, and the number of input channels, middle channels, and output channels of a nested convolution block is defined as follows:

$$\begin{cases} I_{ij} = \begin{cases} f(i), & j = 0 \\ f(i) \cdot j + f(i), & j > 0 \end{cases} \\ M_{ij} = f(i), \quad O_{ij} = f(i) \end{cases} \quad (2)$$

where $I_{ij}$、$M_{ij}$、$O_{ij}$ are the input channels, middle channels, and output channels of the $ij^{\text{th}}$ node, respectively. Note that the middle channels are the output of the first convolution layer, and also the input of the second convolution layer in the nested convolution block. To describe the network structure in detail, we denote the convolution layer, Batch Norm layer, and Rectified Linear Unit [32] as Conv, BN, and ReLu, respectively. The Nested Conv block is a Conv-BN-ReLu with a filter size of $3 \times 3$, padding of 1 and stride of 1, which aims to keep the size of the feature map consistent after each convolution operation. An the last layer of convolution kernel size with $1 \times 1$, the last node of the feature map is mapped into a confidence map using Softmax operation for producing one hot map.

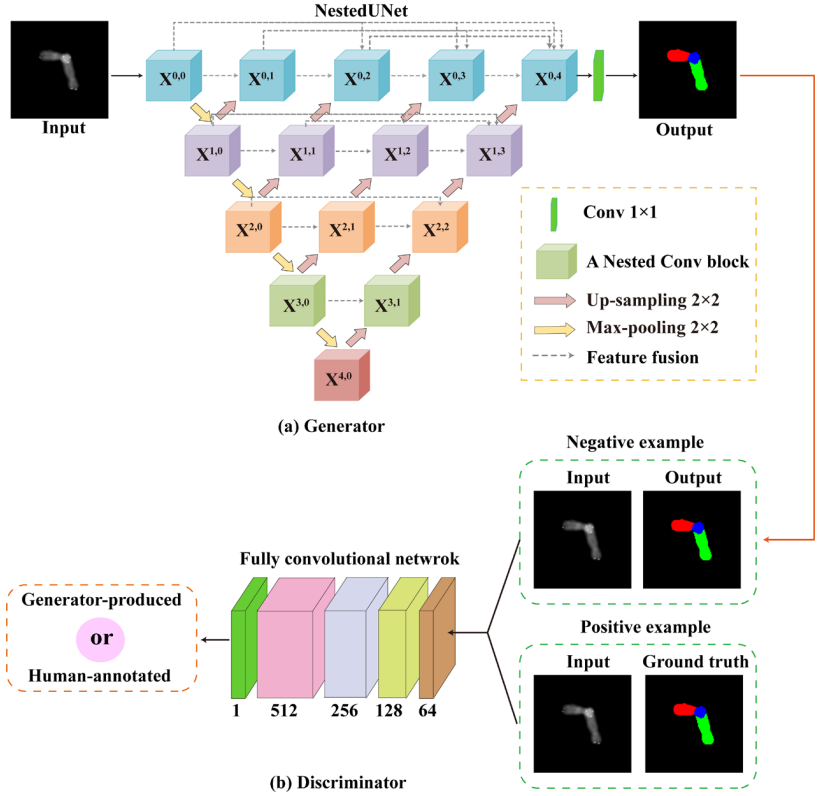

Fig.2. The pipeline of the proposed framework, which consists of two components: a generator and a discriminator. The generator receives a chromosome image as input and outputs a fake segmentation map, and the discriminator attempts to distinguish it from the real ground truth.

*2) Discriminator* As shown in Fig. 2(b), inspired by the PatchGAN in [25], we therefore use a simplified fully convolutional neural network [19, 29] as the discriminator D, which is able to push the outputs distribution closer to the ground truths, making the generator produce high -confident segmentation maps. This discriminator tries to distinguish if each K×K patch in an image is real or fake, and then averages all responses convolutionally across the image to provide the ultimate output of D. Specifically, it consists of five convolution layers with a filter size of 4×4 kernel and {64,128,256,512,1} channels. The first four convolution layers with padding of 2 and the stride of 2, the last two layer with padding of 2 and the stride of 1. Each convolution layer is followed by a Leaky-ReLu parameterized by 0.2 except the last layer. And then a Sigmoid function follow with the last layer, which can produce a binary output for discriminating "real" or "fake" images. Finally, it's worth noting that the input of the discriminator is multi-channel images created by concatenating the source images and the segmented images (generator produced) in the channel dimension, aiming to provide prior information for better-discriminating features. The generator and discriminator are alternately trained using the objective function represented as follows.

### B. Objective Function

1) *Lovász-Softmax* It can optimize the Jaccard index in the continuous optimization framework [26]. Specifically, this method can substantially improve the accuracies of semantic segmentation by optimizing the correct loss during training. Therefore, we choose Lovász-Softmax as the loss of the generator, which can be simplistically defined as:

$$L_{Lova\prime sz\text{-}Softmax} = \frac{1}{C}\sum_{c \in C} \overline{\Delta J_c}(m(c))\overline{\Delta}$$  (3)

where $m(c)$ is a vector of pixel errors for class $c \in C$ aiming to construct the loss surrogate to $\overline{\Delta J}_c$, it's defined by:

$$m(c) = \begin{cases} 1 - f_i(c) & \text{if } c = y_i \\ f_i(c) & \text{otherwise} \end{cases}$$  (4)

In the equation, $y$ is ground truth and $f_i(c)$ is the predicted scores of the model that is mapped to probabilities through a Softmax unit. $\Delta J_c$ is the set function encoding a submodular Jaccard loss for class $c$, indicating a set of mispredictions. Specially, $\overline{\Delta}$ is the surrogate for the minimization of $\Delta$ with first-order continuous optimization, and the elementary operations involved in the calculation of $\Delta$ *(sort)* are differentiable.

2) *GAN loss* First and foremost, we need to choose an appropriate loss function for training our AMFL framework. It is well known that the regular GAN [33] loss is always difficult to convergence and can suffer from model collapsing. We therefore adopt the least-squares generative adversarial networks (LSGAN) [31] as the loss function in our work, which is more stable and can achieve better segmentation results by previous experimental experience [29]. It is defined as:

$$L_{LSGAN}(D) = \mathbb{E}_{i,y \sim P_{data(i,y)}}\Big[(D(i,y)-1)^2\Big] + \mathbb{E}_{i \sim P_{data(i)}}\Big[(D(i,G(i)))^2\Big]$$  (5)

The adversarial learning process is also optimized through the *LSGAN*, which is formulated as:

$$L_{LSGAN}(G) = \mathbb{E}_{i \sim P_{data(i)}}\Big[(D(i,G(i))-1)^2\Big]$$  (6)

Furthermore, in order to make the segmentation map as close as possible to the ground truth, we adopt Lovász-Softmax loss for

supervised segmentation. The objective function for AMFL therefore can be defined as:

$$\min_D L(D) = L_{LSGAN}(D)$$
$$\min_G L(G) = L_{LSGAN}(G) + \lambda L_{Lova\prime sz\text{-}Softmax}$$  (7)

where $\lambda$ controls the relative importance of the two objective functions. Empirically, we set $\lambda$ to 10 in our work.

### C. Evaluation metrics

To quantitatively evaluate the performance of our method, We select eight evaluation metrics including the pixels accuracy (Acc) [34], dice coefficient (Dice) [35], intersection over union (IoU) [34], precision, recall, false-negative rate (FNR), false-positive rate (FPR) [34] and Hausdorff distance (Hausdorff) [34], which are briefly introduced below. For convenience, we use $O$ to denote the output segmentation image, and $G$ to indicate the ground truth. Moreover, the index ranges over the interval [0,1] except Hausdorff.

1) *Acc* It indicates the pixel accuracy of the predicated results in the segmentation. In other words, it represents the proportion of pixels in an image that is correctly predicted. The *Acc* is calculate as:

$$Acc = \frac{\sum_{i=0}^{c} P_{ii}}{\sum_{i=0}^{c}\sum_{j=0}^{c} P_{ij}}$$  (8)

Here $P_{ij}$ means the numbers of pixels that are classified as pixel $j$ but actually belongs to pixel $i$, and $c$ is the categories.

2) *Dice* This metric represents the similarity of predicted image $O$ to the ground truth $G$. The *Dice* is calculated as:

$$Dice_c = 2\frac{|G_c \cap O_c|}{|G_c| + |O_c|}$$  (9)

where $|G|$ and $|O|$ represent the numbers of elements in the arrays.

3) *IoU* This metric represents the intersection area between the predicted image $O$ and the ground truth $G$, the *IoU* can be calculated as follows:

$$IoU_c = \frac{|G_c \cap O_c|}{|G_c \cup O_c|}$$  (10)

4) *Precision* It indicates how reliable the prediction is. This metric can be calculated as:

$$Precsion_c = \frac{TP_c}{TP_c + FP_c}$$  (11)

where $TP_c$ represents the true positives which means the pixels correctly predicted to belong to class $c$, while $FP_c$ represents the false positives, indicating the pixels predicted as class $c$ but do not actually belong to class $c$.

*5) Recall* It indicates how sensitive the prediction is. So, it is also called sensitivity, which can be calculated as follows:

$$Recall_c = \frac{TP_c}{TP_c + FN_c} \qquad (12)$$

where the $FN_c$ represents the false negatives, meaning the pixels predicted as not class $c$ but actually belong to class $c$.

*6) FNR* It's also called the under-segmentation rate, which measures the proportion of the positive classes that is predicted to be negative. It is defined by:

$$FNR_c = \frac{FN_c}{TP_c + FN_c} \qquad (13)$$

*7) FPR* It's also called the over-segmentation rate, measuring the proportion of the negative classes that are predicted to be positive. This metric is calculated as:

$$FPR_c = \frac{FP_c}{FP_c + TN_c} \qquad (14)$$

where $TN_c$ represent the true negatives which mean the pixels that are correctly predicted not to belong to class $c$.

*8) Hausdorff* It represents the shape similarity between the predicted images $\boldsymbol{O}$ and the ground truths $\boldsymbol{G}$. It is calculated by:

$$Hausdorff_c = \max\left\{ \sup_{x \in \boldsymbol{G}_c} \inf_{y \in \boldsymbol{O}_c} d(x, y), \sup_{y \in \boldsymbol{S}_c} \inf_{y \in \boldsymbol{G}_c} d(x, y) \right\} \quad (15)$$

where $d(\cdot)$ represents the Euclidean distance between the pixel points $x$ and $y$. The smaller the Hausdorff distance is, the greater the similarity between the predicted segmentation maps and the ground truth is.

Note that for each metric, a higher value indicates better performance, except for FNR, Hausdorff and FPR, where a lower score gives a better segmentation result.

### D. Baselines and implementation

We validate the effectiveness of our method by comparing it with ten recently state-of-the-art algorithms, including efficient neural network (ENet)[36], bilateral segmentation network version 1 (BiSeNetV1) [37], BiSeNetV2 [38], DeepLabV3+ [39], faster fully convolutional network (FastFCN) [40], U-shape network (UNet) [27], recurrent residual UNet(R2UNet) [41], attention UNet (AttUNet) [42], recurrent residual attention UNet (R2AttUNet) [43], and nested UNet (NestedUNet) [24]. Among them, ENet, BiSeNetV1, and BiSeNetV2 are small-scale models, which usually have smaller network scales and higher inference speed. While the others are large-scale models, which usually have more complex network structures and can learn more potential semantic features. The above methods are used as baselines to evaluate the performance of our method comprehensively.

### E. Selection of the objective function and generator

*1) Selection of the objective function* In order to show the superiority of using Lovász-Softmax as the loss for overlapping chromosome segmentation, as shown in Fig.3, we draw average metric graphs for prevalent losses on all testing sets with all the methods. Intuitively, it can be seen that our method with Lovász-Softmax outperforms all baseline methods in all the metrics. Moreover, it is clear that all baseline methods with Lovász-Softmax also show a leading scoring trend against other losses, indicating that Lovász-Softmax is effective and optimal for overlapping chromosome segmentation feature extraction.

*2) Selection of the generator* In Table I, we evaluate the performance of our method with different generator networks in terms of all the quantitative indicators. Obviously, we can clearly see that the framework with NestedUNet as the generator network is better than other configurations.

## III. EXPERIMENTS AND RESULTS

### A. Preliminary Preparation

*1) Data Preparation and Preprocessing* use the Pommier's overlapping chromosome datasets [44, 45] to verify the effectiveness of the present method. The dataset contains a total number of 13,434 grayscale images with a resolution of $94 \times 93$. For each image, there is a corresponding ground truth, in which each pixel represents an object class. In the segmentation map, class labels of 0, 1, 2 and 3 are denoted as the backgroun

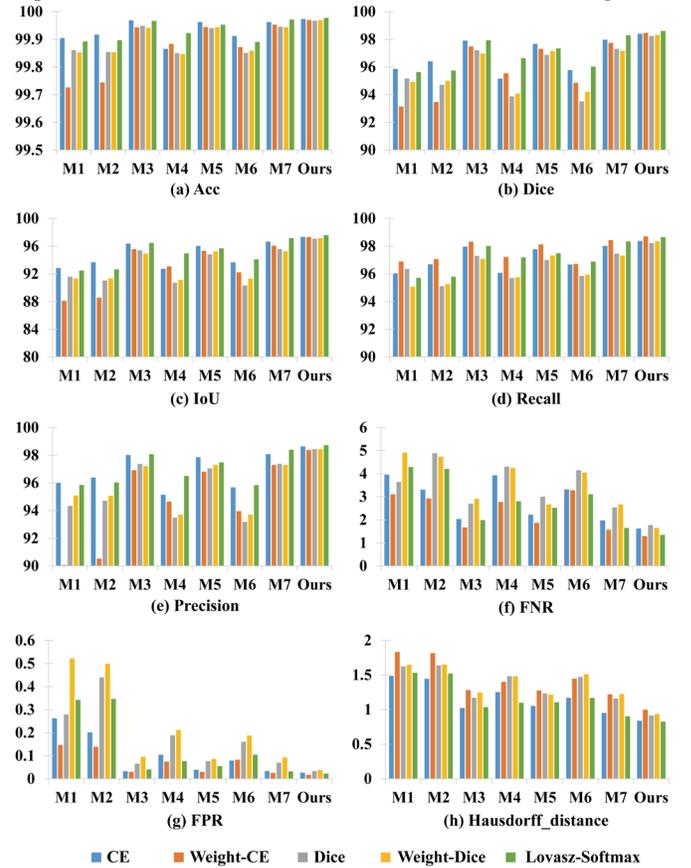

Fig. 3. The average quantitative metrics score of different objective loss functions of various methods. The coordinate scale M1-M7 represent DeepLabV3+, FastFCN, UNet, R2UNet, AttUNet, R2AttUNet and NestedUNet, respectively

TABLE I
THE AVERAGE QUANTITATIVE RESULTS OF PROPOSED METHOD BY USING DIFFERENT GAN GENERATOR

| G | Acc | Dice | IOU | Recall | Precision | FNR | FPR | Hausdorff |
|---|-----|------|-----|--------|-----------|-----|-----|-----------|
| DeepLabV3+ | 99.9149 | 96.4414 | 93.8028 | 96.2027 | 96.9309 | 3.7973 | 0.3094 | 1.4366 |
| FastFCN | 99.9130 | 96.5195 | 93.8791 | 96.5454 | 96.7388 | 3.4546 | 0.2450 | 1.4376 |
| UNet | **99.9727** | 98.2346 | 96.9548 | 98.2836 | 98.3897 | 1.7164 | 0.0314 | 0.9524 |
| R2UNet | 99.9694 | 98.2646 | 97.0231 | **98.6001** | 98.1223 | **1.3999** | 0.0348 | **0.9423** |
| AttUNet | 99.971 | 98.2409 | 96.9582 | 98.2301 | **98.4575** | 1.7669 | **0.0304** | 0.9515 |
| R2AttUNet | 99.9585 | 97.8592 | 96.4989 | 98.1673 | 97.7693 | 1.8327 | 0.0399 | 1.0054 |
| NestedUNet | **99.9776** | **98.6048** | **97.5974** | **98.6550** | **98.7267** | **1.3450** | **0.0227** | **0.8252** |

Note that the units of all indicators are percentages except Hausdorff. A larger value of Acc, Dice, IoU, Recall and Precision indicate a better performance, while a smaller value of FNR, FPR, Hausdorff shows a better performance. The best two results are highlighted in red and green, respectively.

(shown as black), non- overlapping regions of the first chromosome (shown as red), non- overlapping regions of the second chromosome (show as green) and overlapping regions of chromosome (shown as blue), respectively. To match the images with our network, we pad the images to 128×128. The padding value of input images and ground truths are set as 255 and 0 to be consistent with the background of the original images, respectively. We divide the datasets into two subsets: 80% for training (a total number of 10747 images) and the rest 20% for testing (a total of 2686 images), and note that we only kept pairs with ground-truth containing overlapping domains for testing sets (a total of 2432 images)

*2) Implementation* In the training stage, all the training sets are shuffle, and all input images are normalized to the range of 0-1, and the batch size is set to 64. We optimize the generator and the discriminator alternately, both applying the Adam solver with a fixed learning rate of 0.0002, and momentum parameters $\beta_1=0.5$, $\beta_2=0.999$. and then we set random seed to 123.We train our framework from scratch with the training sets to produce the "optimized" model. The training is stopped when training losses did not decrease for 15 consecutive iterations. We save the generator model weights with the best Dice score on the all training sets. For the inference stage, we use the well-trained framework to segment the images. All the experiments are conducted in Pytorch [46] under an Ubuntu OS cloud server with an Intel Xeon(R) CPU E5-2680 v4 @2.40GHz, 40 GB of RAM, and a NVIDIA Tesla P40 GPU with 24 GB of memory. The well-trained models will be available at https://github.com/liyemei/AFML.

### B. Performance

Fig. 4 exhibits some examples of the segmentation results of our method, from which we can see that our method achieves an excellent visual perception result. Moreover, we can also see that the various scales of chromosome individuals and overlapping regions are correctly segmented in all the images, indicating that our method performs well in the multiscale segmentation task. In order to further highlight the superior performance of our algorithm, we show confusion matrices for average accuracy scores on all the testing sets in Fig. 5. We can see that our method shows better results than other state-of-the-art methods. Through carefully comparison, these quantitative results are consistent with the quantitative results in Section

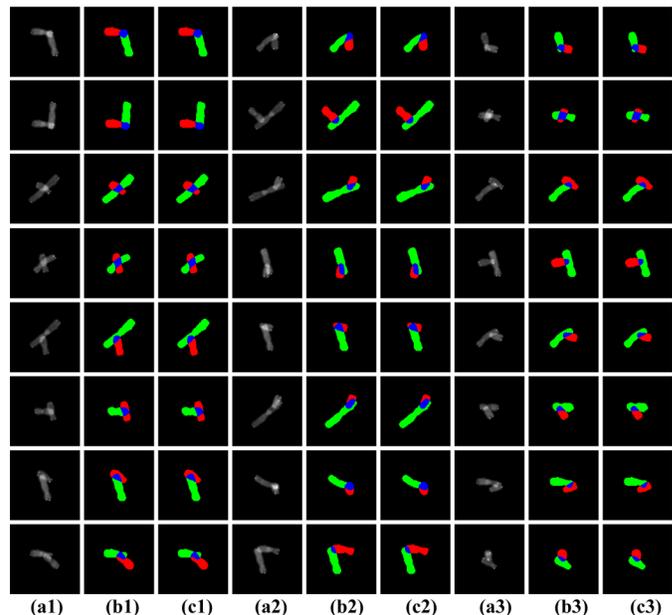

Fig. 4. Some examples of segmentation results of the proposed method. The first to third, fourth to sixth, seventh to nineth columns show the segmentation results of 27 chromosome images, respectively. (a) and (b), (c) are the source images, ground truth and segmentation result, respectively.

III.C, demonstrating the significant superiority of our method, not only for visual perception but also for quantitative analysis.

### C. Performance evaluation

*1) Visual evaluation* In this section, we visually compare the performance of our method with baseline methods. Fig.6 exhibits the results including difference images using pseudo-color map. Here, the difference images are generated through logical multiplication of the inversed ground truth and corresponding predicted result. Fig.6(a)-(j) are the results acquired using baseline methods with CE loss, while Fig.6(k)-(o) are acquired using the presented method with various loss functions. We can see from Fig.6 that our method with Lovász-Softmax or weight-dice loss achieves excellent segmentation results, while the performance of other methods is obviously poor, meaning that these methods do not learn effective features for the overlapping chromosome segmentation. NestedUNet performs the segmentation better than other large-scale models, indicating that multiscale feature learning is helpful for overlapping chromosome segmentation. Furthermore, we can see the difference images acquired with our method is obviously clean than those acquired with other methods, indicating that the cGAN applied in our methods is effective to distinguish the segmented images and ground truths, so as to better learn the features of the chromosomes. Additionally, it is clear that our method with Lovász-Softmax loss segments the images more accurately, where almost every chromosomal region is correctly segmented, compared with other methods. This indicates that the Lovász-Softmax loss helps improve the discrimination ability of our method. In a word, the images shown in Fig. 6 visually show that our AMFL framework can segment overlapping chromosomes with better performance than the baseline methods.

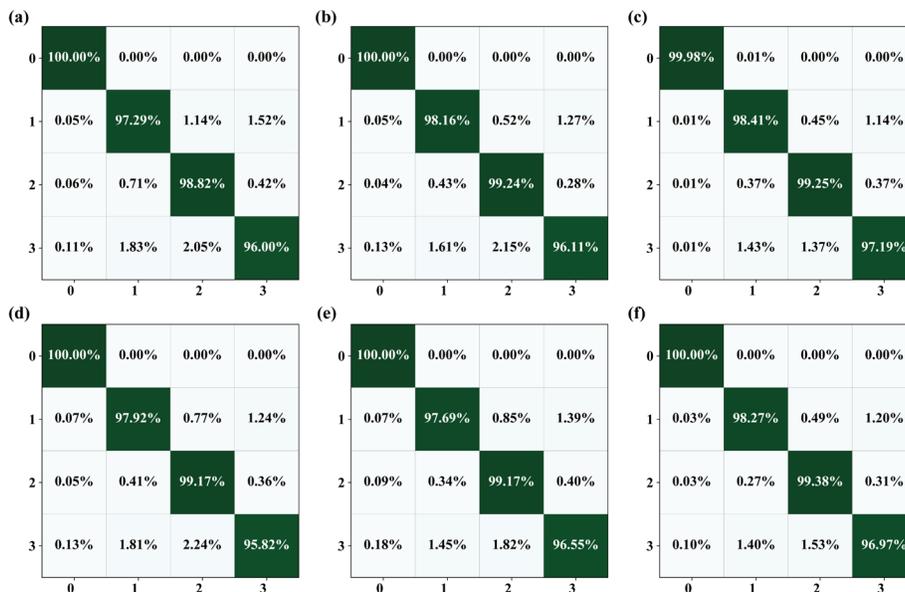

Fig.5. Confusion matrices of average accuracy scores on all the testing sets. Among them, (a) is the average each class IoU score of NestedUNet with CE loss, and (b)-(f) are our method with CE, Weight-CE, Dice, Weight- Dice and Lovasz-Softmax, respectively. For each image, the horizontal axis and the vertical axis represent predicted label and true label, respectively. The coordinate scale 0,1, 2 and 3 represent the background non-overlapping regions of the first chromosome, non-overlapping regions of the second chromosome and overlapping regions of chromosome, respectively. The entry in the *i-th* row and *j-th* column denotes the percentage of the testing samples from class *i* that were classified as class *j*.

2) *Quantitative evaluation* In this part, we quantitatively compare the performance of our method with others and show the results in Table II. Here, we use CE loss for baseline methods. We can see that our AMFL achieves the best performance in all of the metrics. This indicates that using cGAN to discriminate features can push the output distribution closer to the ground truth so that our method outperforms others in overlapping chromosome segmentation tasks. It is also clear that the small-scale models present almost the worst scoring in terms of Dice, IoU and Hausdorff, while large-scale models reach better scores, suggesting that overlapping chromosome segmentation requires a more complex network structure. Again, NestedUNet achieves the top performance than other methods, quantitatively verified the importance of multiscale feature analysis. It is worth emphasizing that our method has a lower Hausdorff distance score, indicating it retains the shape and structure of the chromosome in the output images. The quantitative results, which are consistent with what we can see from Fig. 6 prove the effectiveness of our method in overlapping chromosome segmentation.

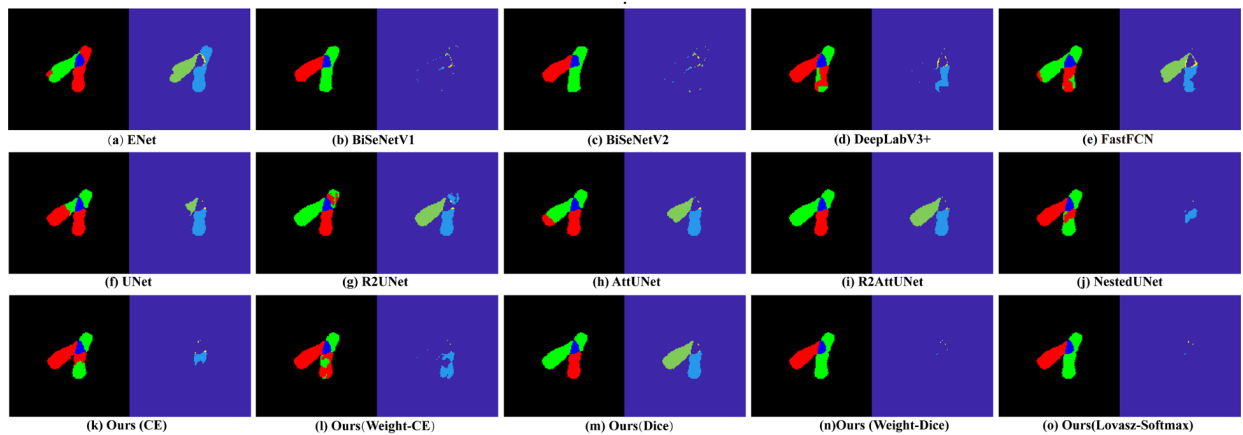

Fig. 6. Examples of the corresponding segmentation results and difference images in pseudo colormap obtained by various methods. The left and right image of each example are the segmentation results and corresponding difference images, respectively. The different colors in pseudo colormap represent incorrect segmentation through the comparison with ground truth.

TABLE II
AVERAGE SCORES OF VARIOUS METHODS ON EIGHT METRICS

| Method | Acc | Dice | IoU | Recall | Precision | FNR | FPR | Hausdorff |
|---|---|---|---|---|---|---|---|---|
| *small models* | | | | | | | | |
| ENet | 99.8707 | 94.5821 | 90.7770 | 94.5365 | 95.0898 | 5.4635 | 0.3791 | 1.5861 |
| BiSeNetV1 | 99.7361 | 90.9037 | 85.1075 | 89.3718 | 93.2404 | 10.6282 | 1.4966 | 1.9584 |
| BiSeNetV2 | 99.8055 | 93.2973 | 88.8068 | 93.1980 | 93.8226 | 6.8020 | 0.6947 | 1.8145 |
| *lager models* | | | | | | | | |
| DeepLabV3+ | 99.9048 | 95.8592 | 92.8454 | 96.0429 | 96.0126 | 3.9571 | 0.2623 | 1.4886 |
| FastFCN | 99.9170 | 96.4061 | 93.6931 | 96.6915 | 96.3868 | 3.3085 | 0.2017 | 1.4452 |
| UNet | 99.9684 | 97.8970 | 96.3765 | 97.9654 | 98.0156 | 2.0346 | 0.0331 | 1.0230 |
| R2UNet | 99.8659 | 95.1638 | 92.7348 | 96.0719 | 95.1458 | 3.9281 | 0.1046 | 1.2535 |
| AttUNet | 99.9625 | 97.678 | 96.0418 | 97.7765 | 97.8637 | 2.2235 | 0.0395 | 1.0528 |
| R2AttUNet | 99.9122 | 95.7752 | 93.6760 | 96.6767 | 95.6791 | 3.3233 | 0.0792 | 1.1688 |
| NestedUNet | 99.9625 | 97.967 | 96.6473 | 98.0266 | 98.0809 | 1.9734 | 0.0341 | 0.9518 |
| **AMFL (Lovász -Softmax)** | 99.9776 | 98.6048 | 97.5974 | 98.655 | 98.7267 | 1.345 | 0.0227 | 0.8252 |

Note that the units of all indicators are percentages except Hausdorff. A larger value of Acc, Dice, IoU, Recall and Precision indicate a better performance, while a smaller value of FNR, FPR, Hausdorff shows a better performance. The best two results are highlighted in red and green, respectively.

Moreover, In order to further highlight the superior performances of our present method. In Table III, we show the average IoU scores of each class and accuracy on all the testing sets, comparing with against methods that are specifically designed for overlapping chromosome segmentation. We can see that our method has a significant advantage over these two methods in terms of the two quantitative metrics score, which proves again the superiority of the AFML framework for overlapping chromosome segmentation. However, due to the imbalance of the categories in the training sets which lack of overlapping areas, resulting in a slightly lower score of average IoU score of Class 3 than Hu et al.'s method, but it's more accord with the diversity of clinical data.

TABLE III
COMPARISON OF AVERAGE IOU AND ACCURACY SCORES FOR EXISTING METHODS

| Method | Average IoU scores | | | Accuracy |
|---|---|---|---|---|
| | Class 1 | Class 2 | Class 3 | |
| Hu et al. [20] [21] | 88.2 | 94.4 | 94.7 | 92.22 |
| Saleh et al. [21] | - | - | - | 99.68 |
| AFML (Ours) | 97.09 | 98.93 | 94.37 | 99.98 |

Note that the index scores for all existing methods are drawn from the references. "_" indicates that it is not described in the paper. And class 1, 2 and 3 are denoted as the two non- overlapping regions of the chromosomes and overlapping regions of chromosome, respectively. The best two results are highlighted in red and green, respectively.

*3) Computational Efficiency* To evaluate the computation

efficiency, we present the total number of model parameters and the average running time of CPU and GPU when using different methods on all the testing sets in Table IV. Although the methods with small-scale networks consume the least resources, obtaining the advancement of rate by sacrificing the accuracy. For the methods with large-scale networks, our method takes about 27ms to segment an image on GPU, which ranks second only behind UNet, and the model parameters are also the second smallest. Our method spends 568ms on CPU, ranking fifth above R2UNet and R2AttUNet. The results show that, in addition to the outstanding segmentation performance, our method also performs well in the computational efficiency, suggesting its great potential in real applications.

*D. Ablation Study*

In order to analyze the role of different parts of the proposed framework, we present the average quantitative results of proposed method by using and without using GAN with different objective function in Table V. Obviously, there are five-fold evidence that can manifest the superiority of our method. First, for the single model without using GAN, we use NestedUNet with Lovász-Softmax achieve improved performance compared with other losses on all the testing sets. Second, the proposed AMFL adopts the GAN mechanism to discriminate features, resulting in achieving a better scoring performance than the individual NestedUNet model without GAN. Third, our method with Lovász-Softmax loss achieves the best performance in most of the metrics. In some cases, it

TABLE IV
THE COMPUTATIONAL EFFICIENCY OF VARIOUS METHODS

| | ENet | BiSeNet V1 | BiSeNet V2 | DeepLab V3+ | FastFCN | UNet | R2UNet | AttUNet | R2AttUNet | NestedUNet | AMFL (Ours) |
|---|---|---|---|---|---|---|---|---|---|---|---|
| Params | 0.35M | 12.43M | 2.85M | 59.46M | 104.3M | 34.53M | 39.09M | 34.88M | 39.44M | 36.63M | 36.63M |
| GPU | 63ms | 19ms | 34ms | 72ms | 78ms | 19ms | 61ms | 27ms | 71ms | 27ms | 27ms |
| CPU | 38ms | 43ms | 34ms | 190ms | 380ms | 274ms | 715ms | 285ms | 734ms | 568ms | 568ms |

**Params:** The total number of model parameters. **GPU/CPU:** Average GPU/CPU runtime measured w.r.t. a full-resolution input (i.e., $128 \times 128$) on all the testing sets.

TABLE V
THE AVERAGE QUANTITATIVE RESULTS OF PROPOSED METHOD BY USING AND WITHOUT USING GAN WITH DIFFERENT OBJECTIVE FUNCTION

| Method | GAN | Acc | Dice | IOU | Recall | Precision | FNR | FPR | Hausdorff |
|---|---|---|---|---|---|---|---|---|---|
| NestedUNet with Dice | × | 99.9457 | 97.3146 | 95.596 | 97.4631 | 97.3846 | 2.5369 | 0.0701 | 1.1617 |
| NestedUNet with Weight-Dice | × | 99.9434 | 97.1637 | 95.2851 | 97.3318 | 97.3021 | 2.6682 | 0.0929 | 1.2242 |
| NestedUNet with CE | × | 99.9625 | 97.9670 | 96.6473 | 98.0266 | 98.0809 | 1.9734 | 0.0341 | 0.9518 |
| NestedUNet Weight-CE | × | 99.9529 | 97.7494 | 96.063 | 98.4318 | 97.3030 | 1.5682 | 0.0256 | 1.2219 |
| NestedUNet with Lovász -Softmax | × | 99.9714 | 98.2898 | 97.1469 | 98.3459 | 98.3993 | 1.6541 | 0.0326 | 0.9036 |
| AMFL with Dice | √ | 99.9670 | 98.2422 | 97.0882 | 98.228 | 98.4479 | 1.772 | 0.0335 | 0.9166 |
| AMFL with Weight-Dice | √ | 99.9699 | 98.3117 | 97.1692 | 98.3524 | 98.4454 | 1.6476 | 0.0381 | 0.9366 |
| AMFL with CE | √ | **99.9735** | 98.4012 | **97.3453** | 98.3788 | **98.6395** | 1.6212 | 0.0266 | **0.8388** |
| AMFL with Weight-CE | √ | 99.9699 | **98.4557** | 97.3259 | **98.7066** | 98.3769 | **1.2934** | **0.0175** | 0.9988 |
| AMFL with Lovász -Softmax | √ | **99.9776** | **98.6048** | **97.5974** | **98.6550** | 98.7267 | **1.3450** | 0.0227 | **0.8252** |

Note that the units of all indicators are percentages except Hausdorff. A larger value of Acc, Dice, IoU, Recall and Precision indicate a better performance, while a smaller value of FNR, FPR, Hausdorff shows a better performance. The best two results are highlighted in red and green, respectively.

gives the second-best performance, demonstrating the effectiveness of using Lovász-Softmax to improve the discrimination ability. Furthermore, we can see that the framework with NestedUNet as the generator network is better than other configurations by observing "Selection of the generator" section, indicating that the modeling combined with multiscale features is effective. Ultimately, we can clearly observe that NestedUNet as the backbone network is better than other competitors, whether using and without using GAN. The above analyses are therefore demonstrating that our AMFL for overlap chromosome segmentation task is effective.

## IV. CONCLUSION AND FUTURE WORK

In this paper, we propose and demonstrate the AMFL framework for overlapping chromosome segmentation. In the network, we adopt NestedUNet to learn multiscale features to detect overlapping regions, utilize cGAN to provide prior information for better discriminating features, employ Lovász-Softmax loss to optimize the segmentation task, and use least square GAN objective to enhance the training stability. To verify the performance of our framework, we compare our method with ten state-of-the-art semantic segmentation methods. The results show that our AFML is superior to others both visually and quantitatively. In the future research, we intend to collect annotated clinical data and design a fully automatic system for the segmentation, classification and karyotype analysis of chromosomes.